%% file: main.tex
\documentclass[11pt]{article}

\usepackage[]{acl}

\usepackage{times}
\usepackage{latexsym}

\usepackage[T1]{fontenc}

\usepackage[utf8]{inputenc}

\usepackage{microtype}

\usepackage[utf8]{inputenc}

\usepackage{tabularx}
\usepackage{tabu}
\usepackage{booktabs}
\usepackage{multirow}

\usepackage{graphicx} 
\usepackage{float} 
\usepackage{subfigure} 
\usepackage{amssymb}
\usepackage{fdsymbol}

\usepackage{hyperref}
\usepackage{tablefootnote}
\usepackage{xspace}

\usepackage{float}
\usepackage{amsmath}
\usepackage{bm}
\usepackage{algorithmicx}
\usepackage{algorithm}
\usepackage{algpseudocode}

\definecolor{amber}{rgb}{1.0, 0.75, 0.0}
\newcommand{\draftonly}[1]{#1} 
\newcommand{\draftcomment}[3]{\draftonly{{\textcolor{#3}{[\textbf{#1--\textsc{#2}}]}}}}
\iffalse
    \newcommand{\yilun}[1]{\textcolor{blue}{}}
    \newcommand{\rui}[1]{}
\else
    \newcommand{\yilun}[1]{\textcolor{blue}{\bf\small [Yilun: #1]}}
    \newcommand{\rui}[1]{\draftcomment{#1}{rui}{amber}}
    
\fi
\newcommand{\ours}{\textsc{ReasTAP}\xspace}
\newcommand{\wikisql}{\textsc{WikiSQL-Weak}\xspace}
\newcommand{\tabfact}{\textsc{TabFact}\xspace}
\newcommand{\wtq}{\textsc{WikiTQ}\xspace}

\newcommand{\logicnlg}{\textsc{LogicNLG}\xspace}
\newcommand{\nreason}{7\xspace}

\newcommand{\tapas}{\textsc{TaPas}\xspace}
\newcommand{\tabert}{\textsc{TaBERT}\xspace}

\title{\ours: Injecting Table Reasoning Skills During Pre-training \\ via Synthetic Reasoning Examples}

\author{Yilun Zhao$^1$ \quad Linyong Nan$^1$ \quad Zhenting Qi$^2$ \quad Rui Zhang$^3$ \quad Dragomir Radev$^1$\\
$^1$Yale University \quad $^2$ Zhejiang University \quad $^3$Penn State University \\
\texttt{\{yilun.zhao, linyong.nan\}@yale.edu}
}

\begin{document}
\maketitle
\begin{abstract}
\input{main/abstract}
\end{abstract}

\input{main/introduction}

\input{main/pre-training}
\input{main/downstream_tasks}
\input{main/experiment}

\input{main/analysis}
\input{main/related_work}

\input{main/conclusion}
\input{main/limitation}
\bibliography{anthology,custom}
\bibliographystyle{acl_natbib}
\clearpage
\appendix
\input{appendix/dataset}
\end{document}

%% file: main/abstract.tex
Reasoning over tabular data requires both table structure understanding and a broad set of table reasoning skills. Current models with table-specific architectures and pre-training methods perform well on understanding table structures, but they still struggle with tasks that require various table reasoning skills.
In this work, we develop \ours to show that high-level table reasoning skills can be injected into models during pre-training without a complex table-specific architecture design.
We define \nreason table reasoning skills, such as numerical operation, temporal comparison, and conjunction.
Each reasoning skill is associated with one example generator, which synthesizes questions over semi-structured tables according to the sampled templates.
We model the table pre-training task as a sequence generation task and pre-train \ours to generate precise answers to the synthetic examples.
\ours is evaluated on four benchmarks covering three downstream tasks including: 1) \wikisql and \wtq for Table Question Answering; 2) \tabfact for Table Fact Verification; and 3) \logicnlg for Faithful Table-to-Text Generation.
Experimental results demonstrate that \ours achieves new state-of-the-art performance on all benchmarks and delivers a significant improvement on low-resource setting.
Our code is publicly available at \url{https://github.com/Yale-LILY/ReasTAP}.

%% file: main/introduction.tex
\section{Introduction}
\begin{figure}[!t]
    \centering
    \includegraphics[width=0.48\textwidth]{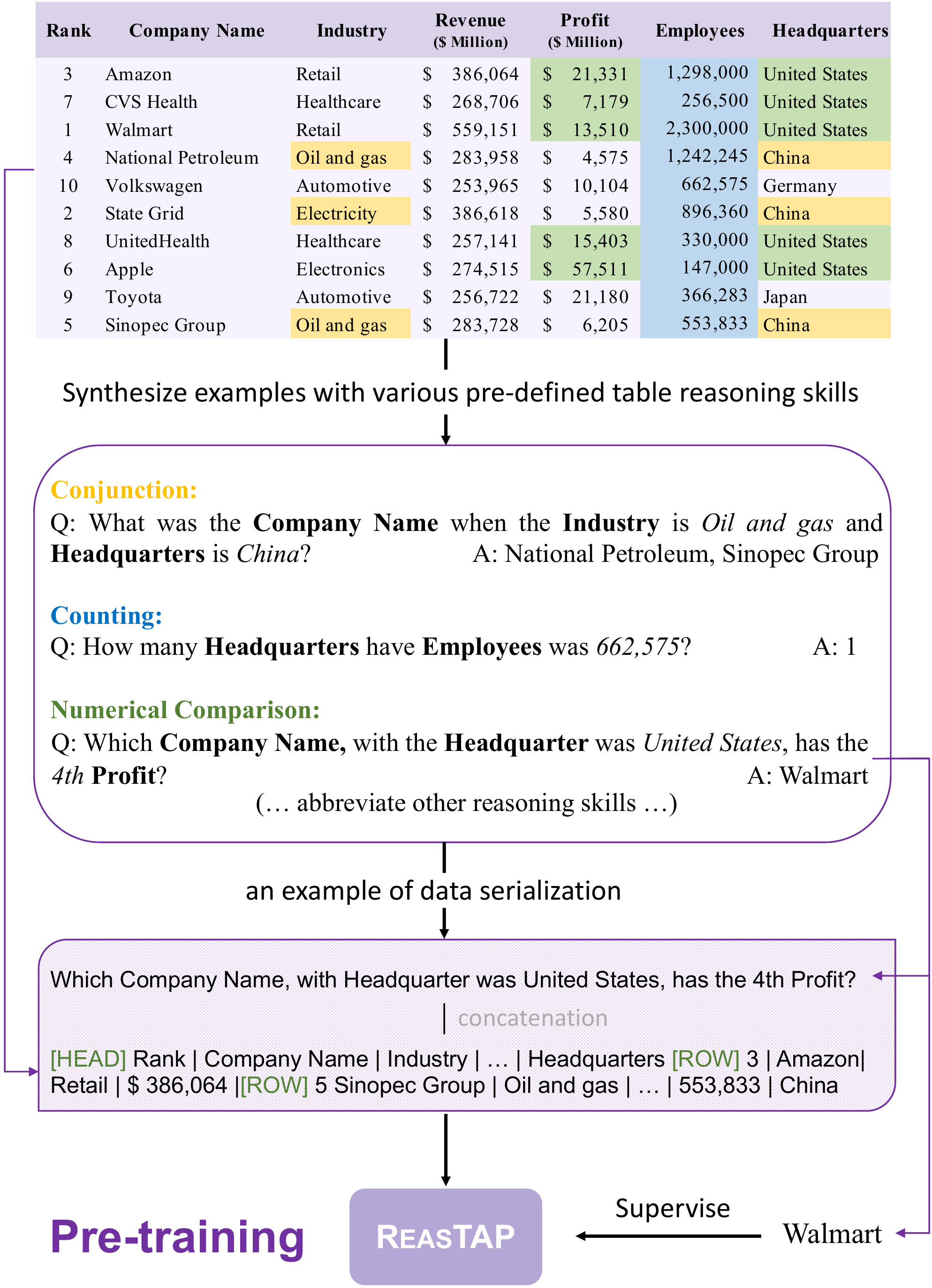}
    \caption{The illustration of \ours pre-training. The tables are crawled from Wikipedia. During pre-processing, we perturb the table row order to alleviate unwanted bias brought by table encoding. The colored cells are relevant facts necessary to answer the given question.  Each color corresponds to a different table reasoning skill. And each reasoning skill corresponds to an example generator, which synthesizes QA pairs over tables according to the sampled templates. We model the pre-training task as a sequence generation task and pre-train \ours to generate correct answers given the flatten table and synthetic question.}
    \label{example}
\end{figure}

Inspired by the massive success of pre-trained language models (LM) on free-form natural language (NL) tasks~\cite{devlin-etal-2019-bert, unilm, 2020t5, lewis-etal-2020-bart}, researchers have attempted to extend the pre-training to table data. Tables are a valuable form of data that organize information in a structured way. They often contain data that is organized in a more accessible manner than in unstructured texts. To adapt the pre-training paradigm on structured tabular data, previous works mainly focus on designing models with table-specific architectures and pre-training methods. This includes introducing a structure-aware attention mechanism~\cite{yin2020tabert, turl, zayats-etal-2021-representations}, adding auxiliary structure indicative embeddings~\cite{tapas1, tapas2, tuta}, and designing table-specific pre-training objectives~\cite{yin2020tabert, yu2021grappa,tuta, liu2022tapex, liu2022plog}.
While these methods are effective in understanding table structures, they increase the modeling complexity and lack interpretability on why models learns table reasoning skills during pre-training.

This paper presents a new table pre-training approach, named \ours, which enables a model to efficiently learn table structure understanding and table reasoning skills during pre-training. 
We first defined \nreason table reasoning skills, such as numerical operation and temporal comparison. As shown in Figure~\ref{example}, for each reasoning skill, a corresponding example generator was applied to synthesize Question Answering (QA) examples over tables.
We modeled the pre-training task as a sequence generation task and pre-trained a sequence-to-sequence (seq2seq) LM to generate the answer to the synthetic questions. \ours is theoretically applicable to any seq2seq LM without a table-specific architecture design. Our key insight is that if a language model can be pre-trained to generate the answers to synthetic questions, which require various table reasoning skills, it should have a great table structure understanding and table reasoning capacity, thereby conferring benefits to downstream tasks. 
The main contributions of our work can be summarized as follows: 
\begin{itemize}
    \item We develop a new table reasoning example generation pipeline, which produces a large-scale table QA corpus that requires various reasoning skills over semi-structured tables.
    \item We propose a new table pre-training method, \ours, which helps the model to learn table structure understanding and various table reasoning skills during pre-training without any table-specific architecture design.
    \item \ours is evaluated on four downstream benchmarks. Experimental results demonstrate that \ours achieves new state-of-the-art results on all of them, and delivers a great improvement on low-resource setting. 
\end{itemize}

%% file: main/pre-training.tex
\section{Pre-training Corpus}
\input{tables/reasoning_example}
\subsection{Table Source and Pre-processing}
We chose publicly available semi-structured tables as the table source.
Specifically, we extracted tables from English Wikipedia\footnote{We parsed the 02-20-2022 Wikipedia dump using WikiExtractor Tools from \url{https://github.com/attardi/wikiextractor}}, which covered a wide range of domains including popular culture, geography, politics, and science. We kept tables with 8-30 rows and at least three columns, resulting in around 600K tables. 
For each extracted table, a pre-processing script was applied to automatically annotate table columns with their data types (i.e, string, number, and date), which allows us to generate questions that involve manipulating numbers and dates. 
Furthermore, recent work~\cite{yang2022tableformer, wang-etal-2022-robust} demonstrates that existing table pre-training approaches might encode table row order as an unwanted bias. 
For example, the pre-trained model being aware of row order information is inclined to select the first or last row of tables when answering superlative-type questions without truly understanding the table content.
To alleviate this problem, we randomly shuffled table rows during pre-processing.

\subsection{Example Generation}
We defined \nreason types of table reasoning skills, with examples and explanations shown in Table~\ref{table:reasoning_example}. The example generation pipeline was adapted from \citet{yoran2021turning}.
Each reasoning skill is associated with one example generator and several question templates. The example generator was implemented as a function that takes a table $T$ and generates several reasoning examples ($T$, $\bm{q}$, $\bm{a}$) according to the template, where $\bm{q}$ denotes the question, and $\bm{a}$ denotes the answer. 

Each template contains typed variables that are instantiated with content from the extracted table. 
Specifically, column \texttt{col} and cell value \texttt{val} are indexed to specify that \texttt{val:i} must be instantiated by a cell value from the \texttt{i}-th column. Some templates also regulate that the selected column and cell value must be date or number type. \texttt{OPERATOR} and \texttt{ORDINAL} correspond to operators and ordinal numerals that are instantiated according to the specific reasoning skill. 
And \texttt{CONDITION:i} can be 1) a cell value from the \texttt{i}-th column; or 2) a number/temporal comparison statement if the \texttt{i}-th column is date or number type.
For example, the question from Figure~\ref{example} "Which Company Name, with Headquarter was United States, has the 4th Profit?" are generated from one of the "Numerical Comparison" templates: "Which \texttt{col:1}, with \texttt{col:2} was \texttt{CONDITION:2}, has the \texttt{ORDINAL} \texttt{col:3}?" 

Once all variables in the sampled template were instantiated, we obtained question $\bm{q}$. Then the example generator would programmatically return the corresponding answer $\bm{a}$.

\subsection{Example Sampling}
After generating a vast number of QA examples for each reasoning skill, we had to sample pre-training data from these synthetic examples. In our setting, the portion of pre-training examples (Table~\ref{table:reasoning_example}) corresponding to each reasoning skill roughly matches the portion of logical operations defined in TabFact~\cite{Chen2020TabFact}. We raised the portion of numerical operation skill as numerical reasoning is more challenging for models to learn. To increase the diversity of pre-training corpus, for each reasoning skill, we also sampled \{SQL query, execution result\} pairs from TAPEX~\cite{liu2022tapex} pre-training corpus as complementary QA examples. The sampled pairs were categorised according to their function (e.g., \texttt{COUNT}, \texttt{SUM}). As a result, we obtained a total of 4M pairs of reasoning examples as the pre-training corpus for \ours.

\section{Pre-training \ours} \label{section:pretrain}
\paragraph{Task Formulation}
Each example in the synthetic pre-training corpus contains a question $\bm{q}$ and a semi-structured table $T$ as the model input. 
The task objective is to generate an accurate answer string $\bm{a}=(a_1, a_2, \dots, a_n)$ given the question $\bm{q}$ and input table $T$:

\begin{equation}
   \boldsymbol{a} = \mathrm{argmax}\prod_{i=1}^n P(a_i | a_{<i}, \bm{q}, T; \theta),
    \label{eq:obj-pt}
\end{equation} 
where $\theta$ denotes the parameters of a seq2seq LM.

\paragraph{Model Architecture}
Our method is theoretically applicable to any seq2seq LM, such as T5~\cite{2020t5} and GPT3~\cite{gpt3}. In our experiments, we implemented \ours based on BART~\cite{lewis-etal-2020-bart}, a widely used Transformer-based pre-trained model~\cite{transformer} that has proved its effectiveness on various comprehension and text generation tasks. 
In our experiments, we chose BART-Large as a backbone, which has around 400M parameters and 12 layers in both encoder and decoder.

\paragraph{Data Serialization}
As illustrated in Figure~\ref{example}, the input contains a question and its corresponding table. We flattened the table so that it can be fed directly into the encore-decoder model. Specifically, by inserting several special tokens to indicate the table boundaries, a flattened table is denoted as 
\begin{equation*}
\begin{aligned}
    T = & \texttt{\small [HEAD]} h_1 ~|~ h_2 ~|~ {\cdots} ~|~ h_m \\
        & \texttt{\small [ROW]} ~~c_{1,1} ~|~ c_{1,2} ~|~ {\cdots} ~|~ c_{1,m} \\
        & \cdots \\
        & \texttt{\small[ROW]} ~~c_{n,1} ~|~ c_{n,2} ~|~ {\cdots} ~|~ c_{n,m}
\end{aligned}
\end{equation*}
where \texttt{[HEAD]} and \texttt{[ROW]} are special tokens indicating the region of table headers and rows respectively. We prefixed the flattened table $T$ with the question and feed them into the model encoder. The decoder is tasked to generate the answer(s), separated by commas, autoregressively.

%% file: tables/reasoning_example.tex
\newcommand{\qm}{\text{\usefont{OT1}{iwona}{m}{n}?}}

\begin{table*}[!t]
\centering
\small
\begin{tabular}{p{1.4cm}p{5cm}p{6.3cm}r}
\toprule
\textbf{Reasoning}  & 
\textbf{Example Templates} & 
\textbf{Example Questions \& Answers} &
\textbf{\%Data} \\
\midrule
    Conjunction &
    What was the \texttt{col:1} when the \texttt{col:2} was \texttt{CONDITION:2} and the \texttt{col:3} was \texttt{CONDITION:3}? &
    \textbf{Q:} What was the \textbf{Television Service} when the \textbf{Country} was \textit{Italy} and the \textbf{Content} was \textit{Sport}? \newline
    \textbf{A:} Sky OMC Sports, ESPN, Gazzetta TV, ... &
    21.6\%
    \\
    \midrule
    
    Quantifiers Only/Every &
    Does \texttt{OPERATOR} \texttt{col:1}, with \texttt{col:2} was \texttt{CONDTION:2}, have \texttt{col:3} \texttt{CONDITION:3}? &
    \textbf{Q:} Does \textit{every} \textbf{Company}, with \textbf{Headquarter} was \textit{Paris}, have \textbf{Industry} \textit{Financials}? \newline
    \textbf{A:} Yes \newline 
    \textbf{Q:} Does \textit{only} \textbf{Company Name}, with \textbf{Founded Year} was \textit{later than 1964}, have \textbf{Employee Number} \textit{greater than 30,000}? \newline
    \textbf{A:} No &
    10.3\%
    \\
    \midrule
    
    Temporal \newline Comparison &
    Which \texttt{col:1}, with \texttt{col:2} was \texttt{CONDITION:2}, happened the \texttt{ORDINAL} according to \texttt{col:3}? 
    &
    \textbf{Q:} Which \textbf{Romaji}, with \textbf{Sales} was \textit{greater than 203,471}, happened the \textit{4th} according to \textbf{Date}? \newline 
    \textbf{A:} Hepburn
    &
    14.5\%
    \\
    \midrule
    
    Date \newline Difference & 
    how much time had passed between when the \texttt{col:1} was \texttt{val:1} and when the \texttt{col:2} was \texttt{val:2}? 
    &
    \textbf{Q:} how much time had passed between when the \textbf{Candidate} was \textit{John Kufuor} and when the \textbf{Candidate} was \textit{Paul McCartney}? \newline
    \textbf{A:} 16 years 
    &
    5.7\%
    \\
    \midrule
    
    Counting &
    How many \texttt{col:1} have \texttt{col:2} \texttt{CONDITION:2}? &
    \textbf{Q:} How many \textbf{Event Location} have \textbf{Attendance} \textit{greater than 10,235}? \newline
    \textbf{A:} 7 
    &
    18.0\%
    \\
    \midrule
    
    Numerical \newline Operation &
    What was the {OPERATOR} of \texttt{col:1} when the \texttt{col:2} was \texttt{CONDITION:2}? &
    \textbf{Q:} What was the \textit{sum} of \textbf{GDP Estimate (\$ US Million)} when the \textbf{GDP Estimate (\$ US Million)} was \textit{greater than 841,969}? \newline
    \textbf{A:}  1,574,013 &
    15.9\%
    \\
    \midrule
    
    Numerical \newline Comparison & 
    Which \texttt{col:1}, with \texttt{col:2} was \texttt{CONDITION:2}, has the \texttt{ORDINAL} \texttt{col:3}? &
    \textbf{Q:} Which \textbf{Franchise}, with \textbf{Owner(s)} was \textit{Nintendo}, has the \textit{5th} \textbf{Total revenue(\$ US Billion)}? \newline 
    \textbf{A:} Pokemon &
    14.0\% \\
    
\bottomrule
\end{tabular}
\caption{\nreason reasoning skills with example for pre-training \ours. Variable names indicate permissible instantiations. \texttt{col} denotes a column name, \texttt{val} denotes a cell value, and indices denote that a cell value must originate from the specified column. \texttt{OPERATOR} and \texttt{ORDINAL} correspond to operators and ordinal numeral that are instantiated according to the specific reasoning skill, e.g., for ‘Temporal Comparison’, \texttt{ORDINAL} is replaced with a reasonable ordinal numeral such as "4th". And \texttt{CONDITION:i} can be 1) a cell value from the \texttt{i}-th column, or 2) number/temporal comparison statement (e.g. "later than 1967") if the \texttt{i}-th column is of number or date type.}
\label{table:reasoning_example}
\end{table*}

%% file: main/downstream_tasks.tex
\input{tables/dataset_overview}
\section{Downstream Tasks}
We evaluated \ours on three different types of downstream tasks to verify its effectiveness. The statistics and examples for each task are shown in Table~\ref{table: downstream}, and Table~\ref{table: downstream_example} in the Appendix, respectively. The fine-tuning of \ours is similar to the procedure for pre-training discussed in section~\ref{section:pretrain}. We modeled both downstream tasks as sequence generation tasks and leverage generative LMs to generate the output autoregressively.

\paragraph{Table QA} \wikisql~\cite{zhongSeq2SQL2017} and \wtq~\cite{pasupat-liang-2015-compositional} were used to evaluate \ours performance on Table QA tasks. \wikisql~\cite{zhongSeq2SQL2017} requires the models to perform filtering and optional aggregation on table cell values to answer the given a question. \wtq~\cite{pasupat-liang-2015-compositional} requires a broader set of reasoning skills, thus is more challenging. The Table QA task formulation is the same as the \ours pre-training task. We used the denotation accuracy, which checks whether the predicted answers are equal to the ground truths, as evaluation metric.

\paragraph{Table Facts Verification}
We chose \tabfact~\cite{Chen2020TabFact} to evaluate \ours performance on Table Facts Verification tasks.
Given a table and a statement, \tabfact~\cite{Chen2020TabFact} tries to distinguish whether the statement is entailed or refuted by the table. \tabfact divides its test sets into Test\textsubscript{simple} and Test\textsubscript{complex} subsets, where Test\textsubscript{complex} contains examples requiring more complex table reasoning skills. 
Furthermore, it selects a small test set Test\textsubscript{small} with 2K samples for human evaluation.
To fine-tune on \tabfact, following BART~\cite{lewis-etal-2020-bart}, we applied a binary classifier upon the hidden state of the last token in the decoder for the output. The objective is to generate the verification label $L \in \{0, 1\}$ given the statement $\bm{s}=(s_1,s_2,{\cdots},s_n)$ and the input table $T$:
\begin{equation}
   L = \underset{i \in \{0, 1\}}{\mathrm{argmax}}\, P(i\,|\, \bm{s}, T\,; \,\theta)
    \label{eq:verification}
\end{equation} 
We used the accuracy (i.e., percentage of correct predictions) as evaluation metric.

\paragraph{Faithful Table-to-Text Generation}
We chose \logicnlg~\cite{chen-etal-2020-logical} to evaluate \ours performance on the Faithful Table-to-Text Generation task. 
Compared with previous Table-to-Text generation benchmarks~\cite{wiseman-etal-2017-challenges, balakrishnan-etal-2019-constrained, parikh-etal-2020-totto, nan2021fetaqa}, which primarily
focus on surface-level realizations without much logical inference, \logicnlg is tasked to generate statements that are logically entailed by the selected table region. Given the serialized input table with its selected columns as $T$, the objective is to generate a sentence $\boldsymbol{y}=(y_1, y_2, \dots, y_n)$ that is both fluent and factually correct:
\begin{equation}
   \boldsymbol{y} = \mathrm{argmax}\prod_{i=1}^n P(y_i | y_{<i}, T\,;\, \theta)
    \label{eq:obj}
\end{equation}
To evaluate the logical fidelity of generated sentences, \citet{chen-etal-2020-logical} proposed two model-based evaluation methods: Parsing-based Evaluation (SP-Acc), and NLI-based Evaluation (NLI-Acc). 
SP-Acc directly extracts the meaning representation from the generated sentence and executes it against the table to verify the correctness. NLI-ACC uses a Natural Language Inference (NLI) model to predict entailment relationships. 
Following \citet{chen-etal-2020-logical}, we used SP-Acc, NLI-Acc as logical-fidelity evaluation metrics; and BLEU-1/2/3 as surface-level evaluation metrics. It is worth noting that higher BLEU scores do not correlate with better logical fidelity~\cite{nan2022r2d2}.

%% file: tables/dataset_overview.tex
\begin{table*}[!t]
	\centering
	\small
	\begin{tabular}{@{}c>{\centering\arraybackslash}p{5cm}cccc@{}}
		\toprule
		\textbf{Task} & \textbf{Dataset} & \textbf{\# Examples} & \textbf{\# Tables} & \textbf{Input} & \textbf{Output} \\
		\midrule
		\multirow{2}*{\textit{Question Answering}} & \wikisql~\cite{zhongSeq2SQL2017} & 80,654 & 24,241 & Question & Answer \\
        & \wtq~\cite{pasupat-liang-2015-compositional} & 22,033 & 2,108 & Question & Answer \\
		\midrule
		\textit{Fact Verification} & \tabfact~\cite{Chen2020TabFact} & 118,275 & 16,573 & Statement & Boolean \\
		\midrule
		\textit{Faithful Text Generation} & LogicNLG~\cite{chen-etal-2020-logical} & 37,015 & 7,392 & Columns & Text \\
		\bottomrule
	\end{tabular}
	
	\caption{Overview of downstream tasks used in this paper.}
	\label{table: downstream}
\end{table*}

%% file: main/experiment.tex
\input{tables/wikisql}
\input{tables/wtq}
\input{tables/tabfact}
\input{tables/logicnlg}

\section{Experiments}
\subsection{Implementation Details}
We implemented our models based on the fairseq library~\cite{fairseq}. We adopted BART-Large as the backbone model.
For table pre-training, we synthesized and sampled 4M pairs of reasoning examples. 
In the following sections, unless specified explicitly, all the experimental results were evaluated under the default settings of 4M reasoning examples and BART-Large configuration.
Our pre-training procedure ran 80,000 steps with a batch size of 256, which took about 34 hours on an 8 NVIDIA A5000 24GB cluster. For downstream tasks, the fine-tuning procedure ran 30,000 steps with a batch size of 256. The best pre-training and fine-tuning checkpoints were both selected according to the validation loss. 

\subsection{Main Results}
\paragraph{Table QA}
On \wikisql, \ours outperforms all the baselines (Table~\ref{table:wikisql}). Specifically, on the test set of \wikisql, \ours achieves a denotation accuracy of 90.4\%, which is 4.3\% higher than BART and 1.2\% higher than the previous best performance. On the more challenging \wtq, as shown in Table~\ref{table:wtq}, \ours also surpasses the previous best system by 1.4\%. 
It is worth noting that compared to \wikisql, \wtq contains much fewer tables and examples, which makes the adaptation of BART to tabular structures more challenging. Further \ours obtains an improvement of 21.1\% over BART, indicating that in the low data regime, the improvements brought by \ours are more significant. We also evaluated \ours performance under low-resource settings (Section \ref{section:low_data}).

\paragraph{Table Fact Verification}
As shown in Table~\ref{table:tabfact}, \ours also obtains a new state-of-the-art accuracy on all test subsets of \tabfact. For example, it surpasses the previous best system by 0.4\% on Test\textsubscript{simple}, and 1.0\% on Test\textsubscript{complex}. 

\paragraph{Faithful Table-to-Text Generation}
Table~\ref{tab:logicnlg-auto} presents the results on \logicnlg. It is observed that the BART backbone has already achieved competitive results in terms of both surface-level and logical-fidelity metrics. Compared with the BART backbone, \ours obtains slightly lower results on BLEU scores, which is reasonable since we continued pre-training \ours on our pre-training corpus that is irrelevant to the text generation task. However, \ours significantly improves the logical-fidelity scores, 
enhancing the SP-Acc and NLI-Acc by 4.7\% and 5.5\%, respectively. The results demonstrate that \ours can also help improve faithful text generation.

%% file: tables/wikisql.tex
\begin{table}[!t]
\centering
\resizebox{0.98\columnwidth}{!}{
\begin{tabular}{lccccc}
\toprule
\textbf{Model}                                & \textbf{Dev} & \textbf{Test} \\
\midrule
\multicolumn{3}{c}{\textit{Previous Models}} \\ 
MAPO~\cite{10.5555/3327546.3327665} & 71.8 & 72.4 \\
MeRL~\cite{pmlr-v97-agarwal19e} & 74.9 & 74.8 \\
LatentAlignment~\cite{wang-etal-2019-learning} & 79.4 & 79.3 \\
HardEM~\cite{min-etal-2019-discrete} & 84.4 & 83.9 \\
\multicolumn{3}{c}{\textit{Pre-trained LMs}} \\ 
TaPas~\cite{tapas1}     & 85.1  & 83.6  \\
GraPPa~\cite{yu2021grappa}  & 85.9 & 84.7 \\
T5-3B~\cite{xie2022unifiedskg} & - & 86.0 \\
TAPEX~\cite{liu2022tapex}  & 89.3  & 89.2 \\
\midrule
BART & 86.9 & 86.1 \\
\ours  & \textbf{91.1} & \textbf{90.4} \\
\bottomrule
\end{tabular}
}
\caption{Denotation accuracies on \wikisql.}
\label{table:wikisql}
\end{table}

%% file: tables/wtq.tex
\begin{table}[!t]
\centering
\resizebox{0.98\columnwidth}{!}{
\begin{tabular}{lccccc}
\toprule
\textbf{Model}                                & \textbf{Dev} & \textbf{Test} \\
\midrule
\multicolumn{3}{c}{\textit{Previous Models}} \\ 
MacroGrammer~\cite{zhang-etal-2017-macro} & 40.6 & 43.7 \\
MAPO~\cite{10.5555/3327546.3327665} & 42.7 & 43.8 \\
MeRL~\cite{pmlr-v97-agarwal19e} & 43.2 & 44.1 \\
LatentAlignment~\cite{wang-etal-2019-learning} & 43.7 & 44.5 \\
IterativeSearch~\cite{dasigi-etal-2019-iterative} & 43.1 & 44.7 \\
\multicolumn{3}{c}{\textit{Pre-trained LMs}} \\ 
T5-3B~\cite{xie2022unifiedskg} & - & 49.3 \\
TaPas~\cite{tapas1}     & 49.9  & 50.4  \\
TableFormer~\cite{yang2022tableformer}             & 51.3 & 52.6 \\
TaBERT~\cite{yin2020tabert}    & 53.0 & 52.3 \\
GraPPa~\cite{yu2021grappa}  & 51.9 & 52.7 \\
TAPEX~\cite{liu2022tapex}  & 58.0  & 57.2 \\
\midrule
BART & 37.1 & 37.5 \\
\ours  & \textbf{58.3} & \textbf{58.6} \\
\bottomrule
\end{tabular}
}
\caption{Denotation accuracies on \wtq.}
\label{table:wtq}
\end{table}

%% file: tables/tabfact.tex
\begin{table*}[!t]
\centering
\resizebox{1.7\columnwidth}{!}{
\begin{tabular}{lccccc}
\toprule
\textbf{Model}                                & \textbf{Dev} & \textbf{Test} & \textbf{Test}\textsubscript{simple} & \textbf{Test}\textsubscript{complex} & \textbf{Test}\textsubscript{small}\\
\midrule
LPA-Ranking~\cite{Chen2020TabFact}            & 65.1  & 65.3 & 78.7 & 58.5 & 68.9   \\
LFC~\cite{zhong-etal-2020-logicalfactchecker} & 71.8 & 71.7 & 85.4 & 65.1 & 74.3         \\
HeterTFV~\cite{yang-etal-2020-program}        & 72.5  & 72.3 & 85.9 & 65.1 & 74.2         \\
SAT~\cite{zhang-etal-2020-table}              & 73.3 & 73.2 & 85.5 & 67.2 & -        \\
TaPas~\cite{tapas1} & 81.0  & 81.0  & 92.3 & 75.6 & 83.9  \\
TableFormer~\cite{yang2022tableformer} & 82.0 & 81.6 & 93.3 & 75.9 & 84.6 \\
DecompTaPas~\cite{yang-zhu-2021-exploring-decomposition}  & 82.7 & 82.7 & 93.6 & 77.4 & 84.7 \\
T5-3B~\cite{xie2022unifiedskg} & - & 83.7 & - & - & - \\
TAPEX~\cite{liu2022tapex}  & 84.2  & 84.0 & 93.7 & 79.1 & 85.5 \\
\midrule
BART & 81.0 & 80.5 & 90.6 & 75.7 & 82.3 \\
\ours  & \textbf{85.1}  & \textbf{84.7} & \textbf{94.1} & \textbf{80.1} & \textbf{86.2}\\
\bottomrule
\end{tabular}
}
\caption{Accuracies on \tabfact dataset.}
\label{table:tabfact}
\end{table*}

%% file: tables/logicnlg.tex
\begin{table*}[!t]
\centering
\resizebox{1.93\columnwidth}{!}{
\begin{tabular}{l|ccc|cc}
\toprule
\multirow{2}{*}{\textbf{Model}} & \multicolumn{3}{c|}{\textbf{Surface-level}} & \multicolumn{2}{c}{\textbf{Logical-fidelity}}\\
\cmidrule(lr){2-4}  \cmidrule(lr){5-6} 
& \textbf{BLEU-1} & \textbf{BLEU-2} & \textbf{BLEU-3} & \textbf{SP-Acc} & \textbf{NLI-Acc} \\
\midrule
GPT2-TabGen~\cite{chen-etal-2020-logical} & 49.6 & 28.2 & 14.2 & 44.7 & 74.6 \\
GPT2-Coarse-to-Fine~\cite{chen-etal-2020-logical} & 49.0 & 28.3 & 14.6 & 45.3 & 76.4 \\
DCVED~\cite{chen-etal-2021-de} & 49.5 & 28.6 & 15.3 & 43.9 & 76.9 \\
T5~\cite{liu2022plog} & 52.6 & 32.6 & \textbf{19.3} & 48.2 & 80.4 \\
PLOG~\cite{liu2022plog} & 51.7 & 32.3 & 18.9 & 48.9 & 85.5 \\
R2D2~\cite{nan2022r2d2} & 51.8 & 32.4 & 18.6 & 50.8 & 85.6\\
\midrule
BART & \textbf{53.0} & \textbf{32.9} & 19.2 & 50.1 & 83.7 \\
\ours & 52.5 & 32.5 & 18.9 & \textbf{54.8} & \textbf{89.2}\\

\bottomrule
\end{tabular}
}
\caption{Performance on \logicnlg test set.}
\label{tab:logicnlg-auto}
\end{table*}

%% file: main/analysis.tex
\section{Analysis}
\input{tables/low_data}
\input{figures/corpus_scale}
\input{tables/necessity}
\input{tables/multi-task}
Experimental results on three different kinds of downstream tasks show that \ours can broadly improve BART's generic table reasoning capabilities, which could be adapted to different downstream tasks, regardless of whether the tasks are highly similar to the \ours pre-training task or not. In this section, we further analyze our approach in terms of various aspects to provide researchers with a deeper insight for future work.
\subsection{Low-resource Setting} \label{section:low_data}
To further understand how well \ours learns table reasoning skills during pre-training, we conducted experiments under the low-resource setting, where we fine-tuned \ours on 20\% and 5\% of downstream task training data. As shown in Table~\ref{table:low_data}, in the low-resource setting, the improvements introduced by \ours are often more significant. For example, with only 5\% training data of downstream tasks, \ours delivers a dramatic improvement of 20.0\%, 21.4\%, and 11.2\% over BART on \wikisql, \wtq, and \tabfact, respectively. The results from the low-resource setting show that \ours endows BART with generic table reasoning capabilities.

\subsection{The Scale of Pre-training Corpus}
Figure~\ref{fig:corpus_scale} illustrates \ours performance on downstream tasks with different pre-training corpus scales. We found that increasing the pre-training corpus generally brings positive effects for all downstream tasks. Furthermore, for simple tasks like \wikisql, the gains by scaling up pre-training corpus are marginal, while for complex tasks like \wtq, it shows a positive trend by scaling up the pre-training corpus. 

\subsection{Necessity of Each Reasoning Skill}
We investigated the contributions of the \nreason reasoning skills to the downstream task performance of \ours. We devised 8 variants of \ours: one was trained with examples from all reasoning skills, while others were trained with examples without one reasoning skill. For each reasoning skill, we sampled 150K examples from the pre-training corpus. We kept the scale of pre-training corpus the same (i.e., 900K). We chose \wtq for experiments, on which BART does not perform well. Results shown in Table~\ref{tab:necessity} demonstrate that all reasoning skills can benefit the model performance on \wtq. Furthermore, we find that some reasoning skills, such as counting and temporal comparison, bring more improvements to the model compared to others.

The analysis also helps us understand how the sets of pre-defined reasoning skills are injected during pre-training. When adopting \ours to a new downstream task that requires new reasoning skills different from existing seven reasoning skills, one can also inject the new reasoning skill into model during the pre-training in a similar way. Specifically, once the templates for the new reasoning skill are designed, the synthesis pipeline will generate new examples for pre-training. Pre-training \ours on these synthetic examples can help model learn the new reasoning skill.

\subsection{Multi-Task Fine-tuning}
We further conducted multi-task fine-tuning experiments to explore whether \ours can benefit from the source task. We chose \wikisql and \tabfact for the source task, as their training datasets are relatively rich, and \wtq as the target task. Models were first fine-tuned on the source task and then fine-tuned on the target task. As shown in Table~\ref{tab:multi-task}, multi-task fine-tuning delivers a significant improvement to the target task when initialized by BART; while the improvements are marginal when initialized by \ours. This is reasonable because most table reasoning skills acquired by multi-task learning have been injected into the model during the pre-training.

%% file: tables/low_data.tex
%

\begin{table*}[!t]
\centering
\resizebox{2.05\columnwidth}{!}{
\begin{tabular}{llllllll}
\toprule
    \multirow{2.5}{*}{\textbf{\%Train}} &
    \multirow{2.5}{*}{\textbf{Model}} &
    \multirow{2.5}{*}{\textbf{\wikisql}} & 
    \multirow{2.5}{*}{\textbf{\wtq}} & 
    \multirow{2.5}{*}{\textbf{\tabfact}} &
    \multicolumn{3}{c}{\textbf{\logicnlg}}\\
    \cmidrule{6-8}
    & & & & & 
    \textbf{BLEU-1/2/3} & 
    \textbf{SP-Acc} &
    \textbf{NLI-Acc} \\

\midrule
\multirow{2.0}{*}{100\%} & BART & 86.9 & 37.1 & 81.0 & 53.2/33.0/19.3 & 51.6 & 84.3\\
& \ours & 91.1 \textbf{(+4.2)} & 58.3 \textbf{(+21.2)} & 85.1 \textbf{(+4.1)} & 52.8/32.7/19.0 & 55.7 \textbf{(+4.1)} & 90.1 \textbf{(+5.8)}\\
\midrule

\multirow{2.0}{*}{20\%} & BART & 76.4 & 21.9 & 77.4 & 52.4/32.5/19.1 & 48.0 & 80.9 \\
& \ours &  86.8 \textbf{(+10.4)} & 47.3 \textbf{(+25.4)} & 82.6 \textbf{(+5.2)} & 52.0/31.9/18.7 & 53.6 \textbf{(+5.6)} & 87.2 \textbf{(+6.3)}\\
\midrule

\multirow{2.0}{*}{5\%} & BART & 61.4 & 14.7 & 63.5 & 48.6/27.1/14.5 & 45.2 & 75.1 \\
& \ours & 81.4 \textbf{(+20.0)} & 36.1 \textbf{(+21.4)} & 74.7 \textbf{(+11.2)} & 47.9/26.3/14.0 & 50.2 \textbf{(+5.0)} & 82.5 \textbf{(+7.4)} \\
\bottomrule

\end{tabular}
}
\caption{Performance on dev set under low-resource setting. Results show the average over 3 random seeds.}
\label{table:low_data}
\end{table*}

%% file: figures/corpus_scale.tex
\begin{figure}[!t]
    \centering
    \includegraphics[width=0.47\textwidth]{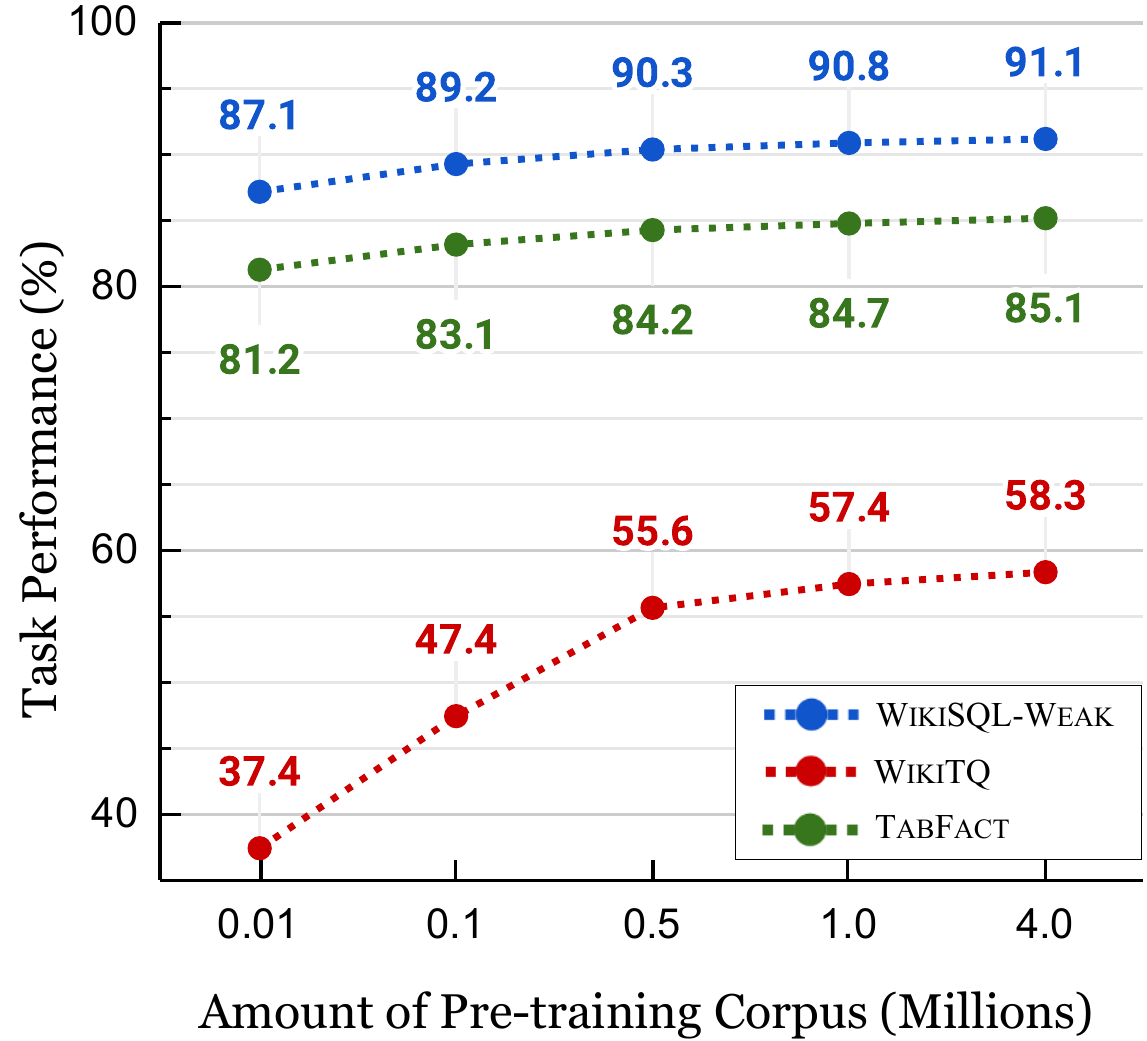}
    \caption{\ours downstream task performance on dev set with different scales of pre-training corpus.}
    \label{fig:corpus_scale}
\end{figure}

%% file: tables/necessity.tex
\begin{table}[!t]
\centering
\resizebox{0.93\columnwidth}{!}{
\begin{tabular}{ll}
\toprule
\textbf{Variant}  & \textbf{Dev}\\
\midrule
BART & 37.1 \\
\textbf{\ours} & \textbf{57.0} \\
\hspace{3mm} w/o Date Difference &  56.7 (-0.3)\\
\hspace{3mm} w/o Quantifiers Only/Every &  56.4 (-0.6)\\
\hspace{3mm} w/o Numerical Operation &  55.6 (-1.4)\\
\hspace{3mm} w/o Conjunction &  55.4 (-1.6)\\
\hspace{3mm} w/o Numerical Comparison &  55.0 (-2.0)\\
\hspace{3mm} w/o Temporal Comparison & 54.8 (-2.2)\\
\hspace{3mm} w/o Counting &  54.3 (-2.7)\\
\bottomrule
\end{tabular}
}
\caption{\wtq dev set denotation accuracy with examples of different reasoning skills removed from the full model pre-training corpus. Each variant is trained using 900K pre-training examples.}
\label{tab:necessity}
\end{table}

%% file: tables/multi-task.tex
\begin{table}[!t]
\centering
\resizebox{0.97\columnwidth}{!}{
\begin{tabular}{lll}
\toprule
\textbf{Source Task} & \textbf{BART} & \textbf{\ours}\\
\midrule
- & 37.1 & 58.3 \\
\tabfact & 42.3 (+5.2) & 58.6 (+0.3)\\
\wikisql & 47.6 (+10.5) & 58.4 (+0.1)\\
\bottomrule
\end{tabular}
}
\caption{Dev denotation accuracy of multi-task fine-tuning on the taget task \wtq.}
\label{tab:multi-task}
\end{table}

%% file: main/related_work.tex
\section{Related Work}
\paragraph{Reasoning Over Tables}
Reasoning over the input context is an important requirement for neural models to be applied in the real world, and especially when the input is structured knowledge such as a table. Several Table QA benchmarks~\cite{pasupat-liang-2015-compositional, zhongSeq2SQL2017, iyyer-etal-2017-search, hybridqa} have been proposed to test systems' capability to conduct different types of reasoning, including numerical, logical or multi-hop reasoning. 
For Table Fact Verification tasks~\cite{Chen2020TabFact, aly2021feverous}, the models are required to perform logical inference to verify whether the given statement is entailed or refuted.
Furthermore, Table-to-Text~\cite{chen-etal-2020-logical, parikh-etal-2020-totto, nan2021fetaqa} tasks to generate a natural language description of some part of the table based on inferences obtained from facts in the contexts. More recently, numerical reasoning over tabular data in financial domain has also raised increasing attention~\cite{tatqa, finqa, zhao2022multihiertt, hitab,li-etal-2022-learning, zhou2022tacube}.

\paragraph{Table Pre-training}
Inspired by the huge success of pre-training in natural language, researchers have attempted to extend pre-training to structured tabular data~\cite{yin2020tabert, tapas1, tapas2, shigap, yu2021grappa, tuta, turl, deng-etal-2021-structure, liu2022tapex} in recent years. Previous table pre-training work such as \tabert~\cite{yin2020tabert} and \tapas~\cite{tapas1, tapas2} 
took corrupted tables and NL sentences as input and tried to recover the corrupted parts. They had the intuition that such recovering processes can help strengthen the linking between sentences and structured tables. On the other hand, TAPEX~\cite{liu2022tapex} learned from synthetic SQL programs. And \citet{jiang-etal-2022-omnitab} further pre-trained TAPEX over natural and synthetic QA examples to improve the few-shot performance over table QA tasks.
Meanwhile, pre-training for Text-to-SQL tasks~\cite{shigap, yu2021grappa, yu2021score, deng-etal-2021-structure} also attracted researchers' attention in recent years. Unlike previous work, we model the pre-training task as a sequence generation task, and inject various table reasoning skills into the model by tasking it to generate the precise answers of reasoning examples.

\paragraph{Synthetic Pre-training Corpus}
Generating a large-scale synthetic pre-training corpus is widely used in both natural language pre-training~\cite{campagna-etal-2020-zero, geva2020injecting,yoran2021turning, neeraja-etal-2021-incorporating, yue-etal-2022-synthetic} and table pre-training~\cite{yu2021grappa,yu2021score,wang2021learning,liu2022tapex}. For example, \citet{geva2020injecting} utilized automatically-generated numerical data to inject numerical reasoning skills during pre-training. And \citet{yoran2021turning} leveraged large-scale WikiPedia resources to automatically generate examples that requires reasoning over multiple facts in the paragraph, and continue pre-training LM on this 
synthetic corpus. Furthermore, recent works~\cite{liu2022tapex, pi2022reasoning} showed that pre-training can be achieved by learning a program executor over synthetic corpus.

%% file: main/conclusion.tex
\section{Conclusion}
In this paper, we propose \ours, a new table pre-training approach, which injects various pre-defined table reasoning skills into models via learning to generate correct answers of synthetic questions. Compared to previous work which design table-specific architectures, \ours is easy to implement and is theoretically applicable to any sequence-to-sequence LM.
\ours is evaluated over four downstream benchmarks.
The experimental results demonstrate that \ours achieves new state-of-the-art results on each of them. 
This includes the improvements on \wikisql denotation accuracy to 90.4\% (+1.2\%); \wtq denotation accuracy to 58.6\% (+1.4\%); \tabfact accuracy to 84.7\% (+0.7\%); and \logicnlg SP-Acc to 54.8\% (+4.0\%), NLI-Acc to 89.2\% (+3.6\%). 
Further analysis demonstrates that \ours delivers a significant improvement to BART on the low-resource setting, indicating that our proposed pre-training approach can effectively improve the model's generic table reasoning capabilities.

%% file: main/limitation.tex
\section*{Limitations}
The main limitation of our approash is that we utilized a template-based method to synthesize pre-training corpus. Although such template-based approach ensures the faithfulness of generated QA examples and the diversity of reasoning process required to answer the questions, it limits the semantic diversity of questions. 
We believe future work could exploit 1) more different types of reasoning skills, such as advanced numerical reasoning skills required in the finance domain~\cite{tatqa, finqa}; 2) a more universal synthetic example generation pipeline; 3) extending models to tables with hierarchical structures (e.g., more than one row or column header)~\cite{hitab, zhao2022multihiertt}; 4) a more efficient training framework ~\cite{biesialska-etal-2020-continual, yoran2021turning} that can update models to learn newly-defined reasoning skills effectively.  

\section*{Ethical Consideration}
Tables used in our synthetic pre-training corpus are collected and extracted from the 02-20-2022 Wikipedia dump\footnote{\url{https://archive.org/details/enwiki-20220220}}, which is publicly available under the Creative Commons Attribution-ShareAlike 3.0 License and the GNU Free Documentation License. The licenses permit us to compose, modify, publish, and distribute additional annotations upon the original content. 

\section*{Acknowledgements}
We would like to thank the anonymous reviewers and action editors for their constructive feedback.

%% file: appendix/dataset.tex
\input{tables/input_example}

%% file: tables/input_example.tex
\begin{table*}[!t]
\small
\centering
\begin{tabular}{p{30mm} p{83mm} p{25mm}}
\toprule
Dataset & Example Input & Example Output  \\ 
\midrule
\wikisql & How many players played for adams state school? \texttt{[HEAD]}Pick \# $|$ CFL team $|$ Player $|$ Position $|$ College \texttt{[ROW]} 145 $|$ calgary stampeders $|$ brett ralph $|$ wr $|$ alberta \texttt{[ROW]} 246 $|$ ottawa renegades $|$ lenard semajuste $|$ fb $|$ adam state $\dots$ & 3 \\ 
\midrule
\wtq & Which coach served previous to ardis smith? \texttt{[HEAD]}Tenure $|$ Coach $|$ Years $|$ Record $|$ Pct. \texttt{[ROW]} 11892 $|$ Shelby Fletcher $|$ 1 $|$ 1–0 $|$ 1.000 \texttt{[ROW]} 21893 $|$ W. M. Walker $|$ 1 $|$ 4–6–1 $|$ .409 $\dots$ & F. C. Owen \\ 
\midrule

\tabfact & John E. Moss and Phillip Burton are both re-elected in the house of representative election. \texttt{[HEAD]}District $|$ Incumbent $|$ Party $|$ Result $|$ Candidates \texttt{[ROW]} Clifornia 3 $|$ John E. Moss $|$ democratic $|$ re-elected $|$ JohnE. Moss (d) 69.9\% John Rakus (r) 30.1\% \texttt{[ROW]} California 5 $|$ Phillip Burton $|$ democratic $|$ re-elected $|$ Phillip Burton (d) 81.8\% Edlo E. Powell (r) 18.2\% $\dots$ & 1 \newline (entailed) \\
\midrule

\logicnlg & Players for Jazz. \texttt{[HEAD]}Player | School / Club Team | Year \texttt{[ROW]} Adrian Dantley | Notre Dame | 1979 - 1986 \texttt{[ROW]} Brad Davis | Maryland | 1982 - 1984$\dots$ & John Duren played for Utah Jazz for 2 years.\\
\bottomrule
\end{tabular}
\caption{
The example inputs and outputs for our model on experimental datasets.
} 
\label{table: downstream_example}
\end{table*}